\documentclass[runningheads]{llncs}

\usepackage{graphicx}
\usepackage{amsmath}
\usepackage{amssymb}
\usepackage[breaklinks=true,bookmarks=false]{hyperref}

\begin{document}

%%%%%%%%% TITLE
\title{
}
\title{Foveated image processing for faster object detection and recognition in embedded systems using deep convolutional neural networks}

\titlerunning{Foveated image processing for object detection and recognition}

\author{Uziel Jaramillo-Avila\\
{\tt\small ujaramilloavila1@sheffield.ac.uk}\\
%\and
Sean R. Anderson\\
{\tt\small s.anderson@sheffield.ac.uk}
}

\authorrunning{U. Jaramillo-Avila and S. R. Anderson}

\institute{Department of Automatic Control and Systems Engineering, University of Sheffield. Sheffield, United Kingdom, S1 3JD}

\maketitle

\begin{abstract}

Object detection and recognition algorithms using deep convolutional neural networks (CNNs) tend to be computationally intensive to implement. This presents a particular challenge for embedded systems, such as mobile robots, where the computational resources tend to be far less than for workstations.  As an alternative to standard, uniformly sampled images, we propose the use of foveated image sampling here to reduce the size of images, which are faster to process in a CNN due to the reduced number of convolution operations. We evaluate object detection and recognition on the Microsoft COCO database, using foveated image sampling at different image sizes, ranging from $416\times416$  to $96\times96$ pixels, on an embedded GPU -- an NVIDIA Jetson TX2 with 256 CUDA cores. The results show that it is possible to achieve a $4\times$ speed-up in frame rates, from 3.59 FPS to 15.24 FPS, using $416\times416$ and $128\times128$ pixel images respectively. For foveated sampling, this image size reduction led to just a small decrease in recall performance in the foveal region, to 92.0\% of the baseline performance with full-sized images, compared to a significant decrease to 50.1\% of baseline recall performance in uniformly sampled images, demonstrating the advantage of foveated sampling.

\end{abstract}

%\begin{IEEEkeywords}
%component, formatting, style, styling, insert
%\end{IEEEkeywords}

\section{Introduction}

%%%%%%%%%%%%%%%%%%%%%%%%%%%%%%%%%%%%%%%%%%%%%%%%%%

Object detection and recognition using deep convolutional neural networks (CNNs) \cite{redmon2016you} \cite{ren2015faster}, has the potential to realise a step change in machine vision for embedded systems, such as in robotics, driverless cars, assistive devices and remote sensors. However, a constant driver for embedded systems is to minimise computational workload to speed up processing and reduce power consumption, noting that the  processing resources for an embedded system tend to be much less than for a workstation. To this end, there has recently been a trend towards developing more compact CNNs for object detection and recognition in embedded systems, which tend to significantly improve the frame rates \cite{shafiee2017fast} \cite{tijtgat2017embedded} \cite{wu2017squeezedet}. 

Computational load in CNN detection-recognition systems can also be reduced by  making the size of the input image itself smaller. The reduction in image size naturally leads to an increase in computational efficiency due to the reduced number of convolution operations in the CNN, but also tends to trade-off against a decrease in detection and recognition performance. Hence, the challenge is to retain detection and recognition performance whilst using small images. The solution that we investigate here is based on foveated image transformation, inspired by photoreceptor density in the human eye.   

% \cite{wong2018tiny}

Human visual perception is dominated by the fovea, a small region of densely clustered photoreceptors in the retina, which accounts for just ${\sim}2\%$ of the visual field \cite{strasburger20
11peripheral}, but as much as ${\sim}50\%$ of the input to neurons in primary visual cortex \cite{wassle1989cortical}. This amplification of the visual field in neural processing is known as cortical magnification factor. In order to see with high acuity, humans actively redirect their fovea towards an object based on saliency (the perceived importance of an object). This active vision system is highly efficient compared to e.g. a passive system with photoreceptors densely distributed throughout the retina, as has been noted elsewhere:
 
% Photoreceptors in the periphery of the retina become sparsely distributed, with density decreasing at an approximately logarithmic rate away from the fovea \cite{}.
%, which is very computationally efficient for analysing visual scenes. As has been noted elsewhere:

\emph{``If the entire 160x175$^{\circ}$ subtended by each retina were populated by photo-receptors at foveal density, the optic nerve would comprise on the order of one billion nerve fibers, compared to about one million fibers in humans.''} \cite{itti2004automatic}

The foveated image processing system in human vision contrasts strongly to how digital images are usually processed in computer vision, where large numbers of pixels are typically used to represent the entire field of view in dense, uniform sampling. Foveated transformation for digital image processing preserves high resolution in the foveal region, centred on an object of interest, whilst compressing the periphery, resulting in reduced image size but no reduction in the field of view. 

There are a number of computational models in use in computer vision to obtain the foveated image, such as the log-polar transform \cite{traver2010review},  the reciprocal wedge-transform \cite{tong1995reciprocal}, and Cartesian foveated geometry \cite{martinez2006new}. The work in \cite{akbas2017object} has demonstrated the advantages of foveated image processing with regard to improvements in computational efficiency (but did not address CNNs).  In recent models of visual saliency using CNNs, images have been applied to networks using a foveal transform  \cite{almeida2017deep} \cite{recasens2018learning}. However, those works did not investigate image size reduction and frame-rate speed-up, which is of critical importance for embedded systems. There is a current gap, therefore, in studying the speed-up effect of foveated transforms on CNNs used for detection and recognition. %\cite{akbas2017object}presents a thorough study of foveated features extraction and pooling for object detection, although predating the predominance of CNNs, which is here addressed, while also using a more complex dataset \cite{lin2014microsoft}.

%\cite{schwartz1977spatial}

The aim of this paper is to investigate quantitatively how detection, recognition and processing speed in a CNN are affected by reducing image size using a foveated transformation. The intention is to demonstrate that foveated image processing coupled with image size reduction enables a  speed-up in processing, whilst retaining high performance in detection and recognition in the foveal region, and reasonable performance in detection and recognition in the periphery. This would provide the foundation for a faster, more computationally efficient object detection and recognition scheme for embedded  systems.

\section{Methods}
\label{methods}

This section presents the key methods used in this paper. In brief, images from the Microsoft COCO database \cite{lin2014microsoft}, were resampled at inreasingly smaller sizes using a foveated transform (Fig. 1), and used to retrain the You Only Look Once version 3 (YOLOv3) CNN for object detection and recognition \cite{redmon2018yolov3}. This foveated approach was compared to linearly downsampled images to analyse the benefit of the foveated transform. To evaluate processing speed, YOLOv3 was implemented on an NVIDIA Jetson TX2 GPU for embedded systems (Fig. 1). In addition, as a comparison against a different type of object detection and recognition system, Faster R-CNN \cite{ren2015faster}, was used with the foveated images to analyse performance (without any retraining) and also compared to YOLOv3 un-retrained. The other advantage of comparing against un-retrained CNNs was that it tests for over-specialisation to the foveated transform in object detection.

\begin{figure} %[t]  %[htbp]
\centerline{\includegraphics[width=1.0\textwidth]{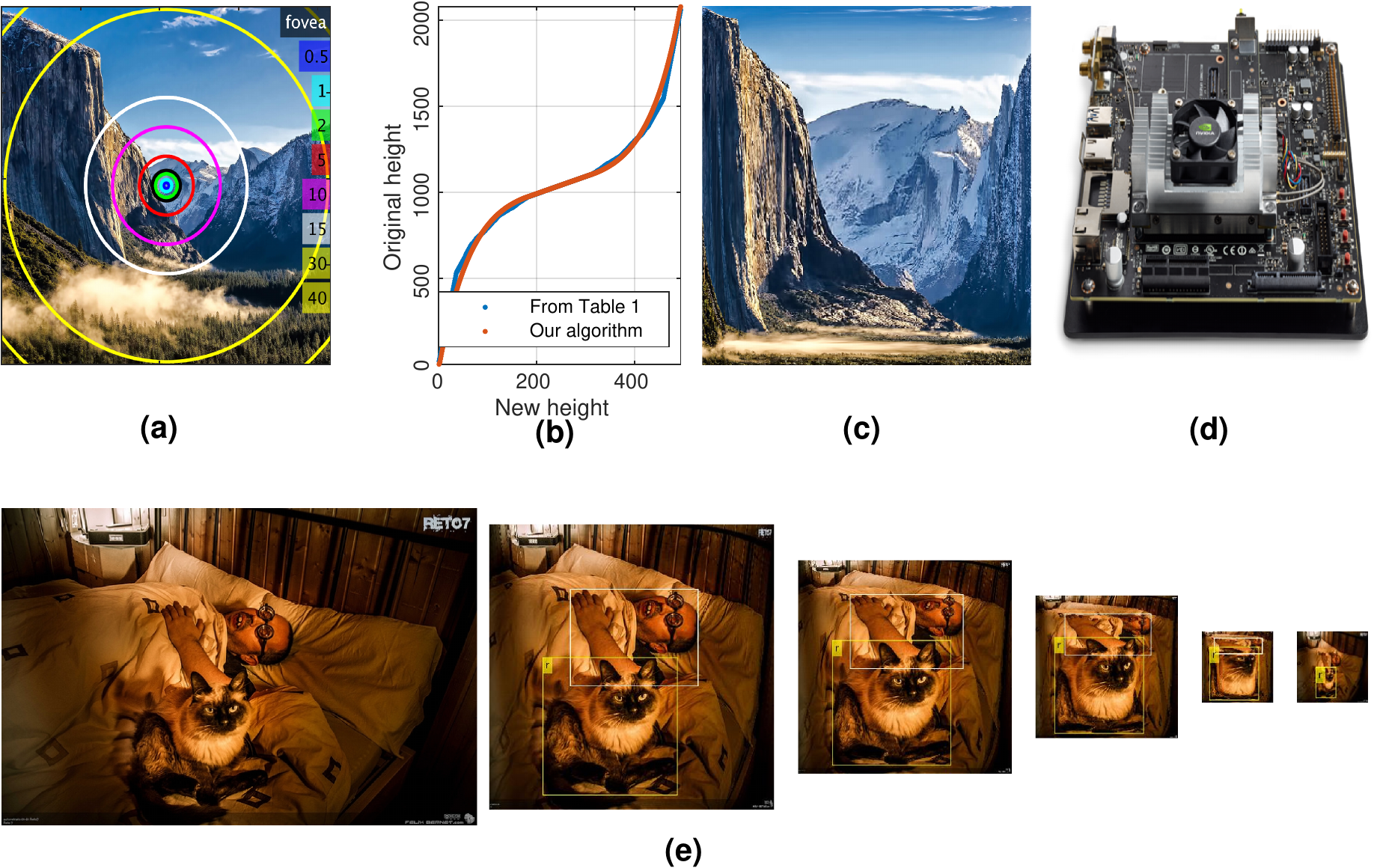}}
\caption{\textbf{(a)} Representation of some key eccentricity values departing from the center of the fovea. 
%The fast decay on the number of data fields is easily visualized comparing the concentric circles with their corresponding value in Table~\ref{table:tab1}; 
\textbf{(b)} Curve representation of the logarithmic behaviour of the rows selected for the example image transformed from 2080 to 494 pixel per side.
%so that it can be transformed from 2080 to 494 pixel per side, assuming that the fovea is in the center of the figure; \textbf{(c)} Resulting foveated image of 494x494 pixels; 
\textbf{(d)} NVIDIA Jetson TX2 for embedded systems (256 Pascal CUDA core GPU, dual-core Denver2 processer, quad-core ARM-A57 processor, 8GB memory, 7.5 Watts, 50$\times$87 mm) mounted on a development board. 
%It is designed to be small-sized embedded computing unit, with power consumption of around 15 Watts at max settings, compared to ranges of 40-250W for personal computers; 
\textbf{(e)} Example of typical image from the COCO dataset used for validation, from left to right, at it's original resolution of 640x426 pixels, at 384x384, 288x288, 192x192, and 96x96, and finally the same image normally downsampled to 96x96 pixels for comparison.}
\label{Resizing_summaryA}
\end{figure}

\subsection{Image Database: Microsoft COCO}
\label{coco_dateset}

The Microsoft COCO  dataset \cite{lin2014microsoft} was used here for training and evaluating the CNNs because it has become a standard benchmark for testing algorithms aimed at scene understanding and pixel-wise segmentation, and it also provides a rich array of relatively context-free images. Note that the COCO database is more challenging than some other standard image databases because the images tend to contain multiple objects as opposed to single objects.

To make the retraining of the CNNs more manageable, a subset of the COCO dataset was taken, considering the first 20 listed objects\footnote{person, bicycle, car, motorbike, aeroplane, bus, train, truck, boat, traffic light, fire hydrant, stop sign, parking meter, bench, bird, cat, dog, horse, sheep, cow}. This dataset was chosen due to the aim of testing the hypothesis in cluttered scenarios in which the ground truth objects are not necessarily centered, having around \emph{3.5 categories and 7.7 instances per image} \cite{lin2014microsoft}. 

The foveated transform (see next section) was applied separately to every object contained in an image in the COCO database. Therefore, a single original image from the COCO database spawned multiple foveated versions of the image, each with the fovea centred on a different object. This increased the number of training images from around 82,000 to 306,000. For retraining the CNNs with uniformly sampled images the number of images was matched to 306,000 by copying each image by the number of objects contained in the image - i.e. to balance the number of uniformly sampled images against the foveated images.

To test and compare performance of uniform image sampling versus foveated image sampling, at different image sizes, the image sizes were varied from 416x416 to 96x96, at intervals of 32 pixels\footnote{96x96, 128x128, 160x160, 192x192, 224x224, 256x256, 288x288, 320x320, 352x352, 384x384 and 416x416}. Note that the upper limit of the image size, 416x416, corresponds to a typical size for running an object detection and recognition algorithm such as YOLO. 

\subsection{Foveal-peripheral image resampling}

%Photoreceptor density and cortical magnification factor has been well studied in the biological domain \cite{schwartz1977spatial, wassle1989cortical, strasburger2011peripheral, wilson1983retino}, which can be used to inform computational models. From a computational point of view, 
A number of different methods have been developed to transform a standard digital image, with uniform sampling, into a foveated image. These include the log-polar transform \cite{traver2010review}, the reciprocal wedge-transform \cite{tong1995reciprocal} and Cartesian foveated geometry \cite{martinez2006new}. As there is no particular consensus on foveated image sampling, the method used here was based on the simple approach of Cartesian log-spaced sampling, which captures the key feature of densely sampling the fovea and compressing the periphery. This method was found to be effective, and because it distorts the original uniformly-sampled image less than, e.g. a log-polar transform, it gives the additional key benefit of enabling the use of transfer learning to speed-up the training of the CNNs (i.e. initialising the CNN weights using a network pre-trained on uniformly sampled images).

The basic approach we take is to resample the uniform digital image of size $N_x \times N_y$ pixels, to a new size of $n_x \times n_y$ pixels with log-spacing, so that for the upper right quadrant of the image with the fovea centred on $(x_0,y_0)$ we have sample locations,
\begin{gather}
    x_k= \exp{(k \Delta_x)} \quad \text{  for  } k=0,\ldots,n_x/2 \\
    y_k= \exp{(k \Delta_y)} \quad \text{  for  } k=0,\ldots,n_y/2
\end{gather}
where 
\begin{gather}
    \Delta_x= 2 n_x^{-1} \log{(N_x/2)} \\
    \Delta_y= 2 n_y^{-1} \log{(N_y/2)}
\end{gather}

To illustrate the performance of this model, a uniformly sampled image of $2080 \times 2080$ pixels is shown in Fig.~\ref{Resizing_summaryA}(a): if we use the eccentricity/data field values in Table~\ref{table:tab1} from \cite{wilson1983retino}, we observe that the sampling model given above fits well to this data Fig.~\ref{Resizing_summaryA}(b), and produces the foveated image of $494 \times 494$ pixels shown in Fig.~\ref{Resizing_summaryA}(c).

To illustrate the foveal sampling algorithm an example of an image downsampled logarithmically, at different resolutions, can be found in Fig.~\ref{Resizing_summaryA}(e), where the fovea is focused in a cat in the lower half of the image. 

% \begin{equation}
% x_a' = \log x_a \in \mathbb{Z}, [1, x_c]  \label{Xi_left}
% \end{equation}
% \begin{equation}
% x_b' = \log x_b \in \mathbb{Z}, [x_c, w]  \label{Xi_right}
% \end{equation}
% \begin{equation}
% x' = x_a' \cup x_b' \quad \text{and} \quad y' = y_a' \cup y_b'
% \end{equation} 
% \begin{equation}
% A_{w',h'} = \begin{matrix}
% I_{x_1' y_1'} & \ldots & I_{w' y_1'}\\
% \vdots & \ddots & \vdots\\
% I_{x_1' h'} & \ldots & I_{w' h'}
% \end{matrix}
% \end{equation} 

% To downsample an image I of width $w$ with the fovea centred in $x_c$ in the x axis, where it is a proportion (from 0 to 1) to $w$, to a desired size $w'$, a matrix $x_a'$ of size $(w' + C)*x_c + w*x_c + 1$ is created, within range 0 to $w*x_c$. C is a integer counter that iteratively increases value. $x_a'$ represents the pixel index columns chosen to the left of the fovea, so a second matrix $x_b'$ is required, from $w*x_c$ to $w$ of size $(w' + C)*(1 - x_c) + w*x_c + 1$. After taking $I_x = x_a' \cup x_b'$ and eliminating the repeated elements of $I_x$, it consistently ends up with a smaller number elements than desired, then the value of $C$ is increased; $C = C + 1$, and the process is repeated. 

\begin{table*}[htbp]
     \begin{center}
         \begin{tabular}{|c|c|c|c|c|c|c|c|c|c|c|}
         \hline
         \textbf{Eccentricity} & 0.5 & 1 & 2 & 5 & 10 & 30 & 45 & 60 & 70 & 90\\
         \hline
         \textbf{Number of data fields} & 256 & 552 & 848 & 1239 & 1534 & 2003 & 2176 & 2299 & 2365 & 2472\\
         \hline

         \end{tabular}
     \end{center}
 \caption{Relationship between the eccentricity angle in the eye and the number of data fields \cite{wilson1983retino}, where they represent retinal regions over which stimulus is collected in cell sub-assemblies from thousands of input fibers and overall properties are calculated over them. Data reproduced from \cite{wilson1983retino}.}
 \label{table:tab1}
 \end{table*}

%\begin{figure}
%\lstinputlisting[language=Octave]{pseudocode1.m}        % include pseudocode 
%\caption{Pseudocode test}
%\label{Pseudocode1}
%\end{figure}

%The main aim here pursued is to evidence the benefit of using a fovea like sensorial input, rather than to propose a new adaptation to the current vast collection of network architectures available. For this reason, it was opted to test on two state-of-the-art models; \emph{You Only Look Once} \cite{redmon2016you} and \emph{Faster R-CNN} \cite{ren2015faster}, both are widely used and freely accessible, and where tested as implemented in their official repository and Pytorch \cite{paszke2017automatic}, respectively. 

%\begin{equation}
%Y(x) = 5.218 \cdot 10^{-5} x^3 - 0.03875 x^2 + 10.06 x + 131.7   \label{eq1}
%\end{equation}

%\begin{equation}
%Y(x) = 5.978 \cdot 10^{-5} x^3 - 0.04439 x^2 + 11.32 x + 50.44   \label{eq2}
%\end{equation}

\subsection{Object recognition with YOLO}

The foveated image processing method was tested and evaluated here using the object detection and recognition algorithm based on You Only Look Once (YOLO) \cite{redmon2016you}, specifically using the current latest version of the algorithm (YOLOv3) known as Darknet-53 \cite{redmon2018yolov3}, as found in its original repository \cite{darknet13}. This algorithm was selected because it has become well established since its inception.

In brief, the YOLOv3 network is a CNN with 53 layers (hence the label Darknet-53). The network is designed with successive blocks, where each block is composed of a $1\times1$ convolutional layer, followed by a $3\times3$ convolutional layer, and a residual layer. Blocks are repeated numerous times with occasional shortcut connections, followed by average pooling then a fully connected layer with softmax output.

YOLOv3 uses dimension clusters as anchor boxes to predict the object bounding boxes along with the class label \cite{redmon2018yolov3}. The system outputs 4 coordinates to define each bounding box:  the centre coordinates of the box $(x,y)$, the width, $w$, and height, $h$. The loss function for each of these variables is defined as the sum-of-squared error. The class label prediction is done for the objects contained in each bounding box using multilabel classification, which is trained using a binary cross-entropy loss function.

%Given the high time and computational requirements for training a network in a bast dataset such as COCO, it proves hard to exhaustively test on all new architectures that are added to the literature, specially at the speed in which it evolves. We aim at using two models representative and at a competitive level; while Faster R-CNN has a dedicated convolutional \emph{Region Proposal Network}, YOLO is a regression model which penalizes the sum of errors for inaccurate bounding boxes, the center of the cell, object confidence and class prediction. 

%with the \emph{loss function} shown in Eq.\ref{yolo_loss}:   

%\begin{equation}

%\lambda_{coord} \sum_{i=0}^{S^2}\sum_{j=0}^B \mathbbm{1}_{ij}^{obj}[(x_i-\hat{x}_i)^2 + (y_i-\hat{y}_i)^2 ] \\&+ \lambda_{coord} \sum_{i=0}^{S^2}\sum_{j=0}^B \mathbbm{1}_{ij}^{obj}[(\sqrt{w_i}-\sqrt{\hat{w}_i})^2 +(\sqrt{h_i}-\sqrt{\hat{h}_i})^2 ]\\ &+ \sum_{i=0}^{S^2}\sum_{j=0}^B \mathbbm{1}_{ij}^{obj}(C_i - \hat{C}_i)^2 + \lambda_{noobj}\sum_{i=0}^{S^2}\sum_{j=0}^B \mathbbm{1}_{ij}^{noobj}(C_i - \hat{C}_i)^2 \\ &+ \sum_{i=0}^{S^2} \mathbbm{1}_{i}^{obj}\sum_{c \in classes}(p_i(c) - \hat{p}_i(c))^2 

%\label{yolo_loss}
%\end{equation}

Training was performed here on an NVIDIA GeForce GTX 1070 GPU (with 1,920 Pascal CUDA cores), using 306,000 images (derived from the COCO database), with a batch size of 32, and subdivision of 16. Stochastic gradient descent with momentum was used as the training algorithm, with learning rate of 0.001, momentum of 0.9 and weight decay of 0.0005. Network weights were initialised using the pre-trained network obtained from \cite{darknet13}. The training process was iterated for 200,000 steps ($\sim9,500$ iterations per epoch, i.e. for 20 epochs), repeating the entire process for the 11 image sizes listed in section \ref{coco_dateset}, where each CNN was restructured to match the size of the input images.  Testing was done on a reserved validation data set of 6000 images.

Frame-rate was evaluated by processing all test images and averaging the result on an NVIDIA Jetson TX2 board, with a 256-core Pascal GPU, using CUDA Toolkit 8.0 and cuDNN 5.1, with the images saved on internal memory but ignoring the time required to load or display them. 

\subsection{Object recognition with Faster R-CNN}

Faster-RCNN is an object detection and recognition system that uses a region proposal network (RPN) to generate region proposals for object detection and recognition by a subsequent CNN \cite{ren2015faster}. Faster-RCNN is particularly efficient because the RPN shares convolutional features with the detection/recognition CNN. In this paper, Faster-RCNN is used without retraining as a comparison to YOLO for processing foveated images. The specific implementation used is the current version developed by the original authors \cite{paszke2017automatic}.

%A considerable focus of innovation of faster R-CNN is the way in which region proposals are generated with, by default, 9 anchor boxes for every position of the sliding window. It works by a Region Proposal Network which is trained with a loss function evaluating the probability of a box either being an object or not \cite{ren2015faster}. This models was tested as implemented in Pytorch \cite{paszke2017automatic} by their original contributors.

\subsection{Analysis and Evaluation}

Performance is evaluated here using the averaged precision and recall metrics \cite{sokolova2009systematic},
%
%\begin{equation}
%\text{Precision} = \frac{\sum_{i=1}^n \text{TP}_i}{\sum_{i=1}^n (\text{TP}_i + \text{FP}_i)}  %\label{precision_eq}
%\end{equation}
%
%\begin{equation}
%\text{Recall} = \frac{\sum_{i=1}^n \text{TP}_i}{\sum_{i=1}^n (\text{TP}_i + \text{FN}_i)}  \label{Recall_eq}
%\end{equation}
\begin{equation*}
\text{Precision} = \frac{\sum_{i=1}^n \text{TP}_i}{\sum_{i=1}^n (\text{TP}_i + \text{FP}_i)}
\quad\mathrm{and}\quad
\text{Recall} = \frac{\sum_{i=1}^n \text{TP}_i}{\sum_{i=1}^n (\text{TP}_i + \text{FN}_i)}  
\end{equation*}
where $\text{TP}$ are the true positives, $\text{FP}$ are the false positives, and $\text{FN}$ are the false negatives. A true positive is only counted if the CNN predicts the correct class label \emph{and} the object location, as measured by an Intersection over Union (IoU) value \cite{redmon2016you}, is over a threshold, set to $0.5$ as in \cite{redmon2016you} \cite{redmon2018yolov3}, a ratio of the between the area of overlap and the area of union between the prediction and the ground-truth. It is important to differentiate that, while it is assumed that the fovea is centered in the object, object localization still needs to be evaluated as passing the bounding box IoU threshold.

%, i.e. this requires that there is a $50\%$ overlap of the predicted bounding box with the ground-truth box for the prediction to be considered a match. 

%Since a very precise location would prove difficult to be set in the periphery, but for an autonomous agent, such as a robot, it is easy to switch the fovea to a new region of interest. Following the same logic, only the top-1 prediction is taken, since it is the one likely to be considered by a robot to make autonomous decisions. 

Performance is evaluated separately in the fovea and the fovea-periphery (to explicitly quantify performance in the foveal region where detection-recognition should be accurate, and in the periphery where we expect accuracy to decrease). 

\section{Results}

To recap, the aim of this paper was to analyse the object detection and recognition performance of CNNs using foveated images, with reduced image size to enable speed-up in processing. Varying image sizes were evaluated along with two CNNs: YOLOv3 and Faster-RCNN. The CNN frame rates were analysed from implementation on an NVIDIA Jetson TX2 -- a GPU designed for embedded systems.

The results are analysed in this section in three parts: 1. foveal performance on retrained YOLOv3 (with 20 object classes); 2. foveal-peripheral performance on retrained YOLOv3 (with 20 object classes); 3. foveal and foveal-peripheral performance on un-retrained Faster-RCNN versus un-retrained YOLOv3 (both with 80 object classes).  Comparing against un-retrained systems also tests for over-specialisation to the foveated transform itself in object detection.

\subsection{Foveal analysis on YOLO with re-training}
\label{foveal_analysis}

The foveal analysis performed in this section assumes that a saliency step has already been performed that crudely aligns the fovea with a point of interest. The CNN still has to detect the object precisely, in terms of the bounding box, and also recognise the object using classification.

\begin{figure}
\centerline{\includegraphics{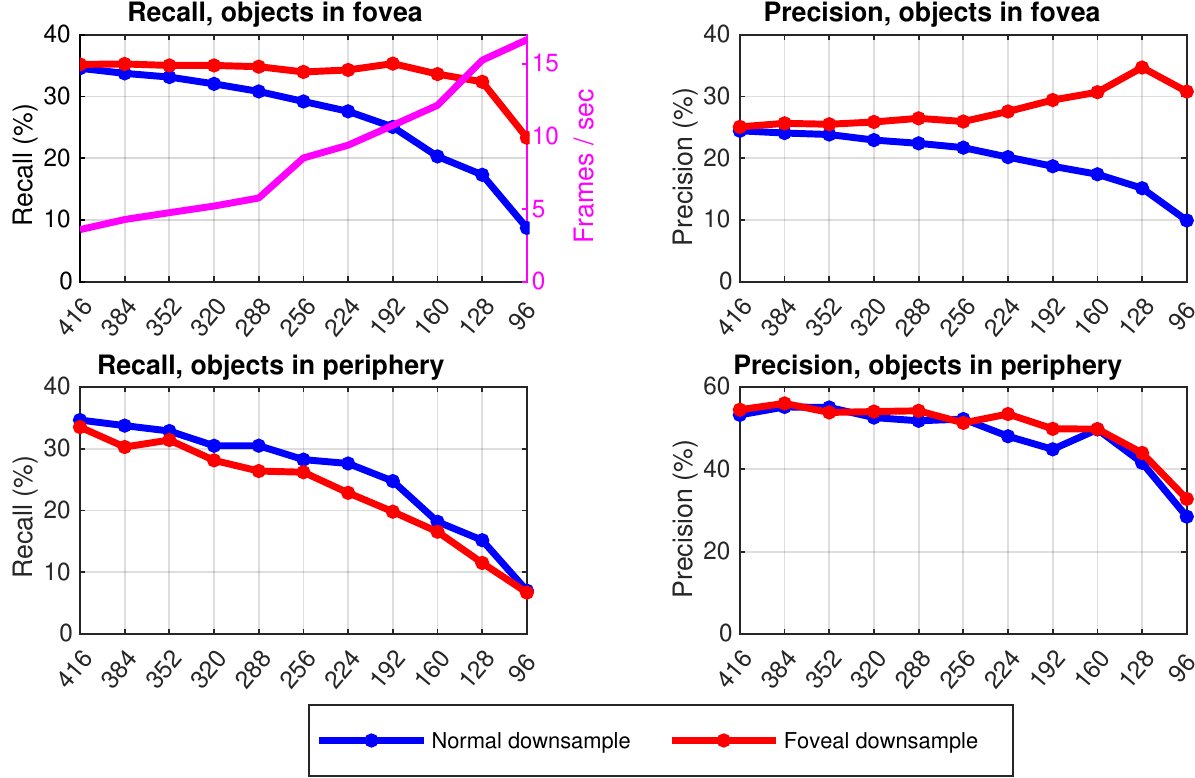}}
\caption{Recall and precision performance for the YOLO network trained at different resolutions. Top row: objects in the fovea. Bottom row: objects in the periphery. Top left also shows the average frame rate for processing each image size.
%The top graph highlights the assertion that, with the object in the fovea, the performance rate can be kept relatively constant (35.20\% at 416x416 to 32.38\% at 128x128). This behaviour is contrasted, in the right-Y axis, with the average frame-rates that were achieved in a Jetson TX2 board. In the top right, the increase in precision for using foveal images is notable. %This boost can be expected since the object of interest takes a big portion of the scene, but most remarkably the algorithm learns to make less false positive predictions. 
%The bottom right quadrant shows the performance for all objects, with an accuracy decrease notably similar to that observed for normal downsampling. Some of the key results are presented numerically in Table~\ref{table:retResults}.
}
\label{Coco20RetrainedA}
\end{figure}

Baseline recall performance of YOLO using foveated and uniformly sampled images at the largest image size tested, $416 \times 416$ pixels, was similar at 35.20\% and 34.57\% respectively (Fig.~\ref{Coco20RetrainedA} and Table~\ref{table:retResults}). The framerate on the Jetson TX2 at this image size was just 3.59 FPS. 

The recall at size $128 \times 128$ using foveated images decreased only slightly to 32.38\% (92.0\% of the baseline result) but for uniformly sampled images decreased to $17.33\%$ (50.1\% of the baseline result) - Table \ref{table:fovTable}. The key additional point is that frame rate increased to 15.24 FPS at an image size of $128\times 128$ - this is over a $4\times$ speed-up.

Interestingly, the precision performance increased for the foveated images as image size reduced, but decreased for the uniformly sampled images (Fig.~\ref{Coco20RetrainedA} and Table~\ref{table:retResults}). The increase in precision performance for foveated images is likely a benefit of the fact that the object of interest takes up more of the visual scene, reducing the false positives.

%\begin{figure}
%\centerline{\includegraphics[width=1.0\textwidth]{Coco20Learning_evA}}
%\caption{\textbf{(a)} Recall performance evolution for foveated images, through different epochs of training. \textbf{(b)} Recall performance evolution for normally downsampled images after 20 epochs.}
%\label{Learning_ev}
%\end{figure}

%\textbf{(a)} Illustration of the rows and columns selected to transform an image of original size 640x360 to 100x100, centering the fovea in 416x162 (0.65, 0.45). The red rectangle represents the borders of an image in the fovea. The magenta rectangles represent different "objects in the periphery".

\subsection{Foveal-peripheral analysis on YOLO with re-training}
\label{peripheral_analysis}

The foveal-peripheral recall performance was similar to the foveal-only performance at the largest image size, $416 \times 416$ pixels, for foveated images (35.20\%) and uniformly sampled images (34.57\%) (Fig.~\ref{Coco20RetrainedA}). As image size was reduced to $128 \times 128$ pixels, recall performance for both foveal and uniformly sampled images decreased significantly, to 34.3\% of baseline for foveated images and 45.3\% of baseline for uniform images (Fig.~\ref{Coco20RetrainedA} and Table \ref{table:perTable}).  The decrease in precision is less pronounced across the same range (to about $\sim$30\% for both image types), indicating that precision is less sensitive to the reduction in image size in the periphery. The precision-recall curve for the smaller networks is shown in Fig. \ref{Recall_precision}.

\begin{table*}
    \begin{center}
        \begin{tabular}{|c|c|c|c|c|c|c|c|c|c|c|c|}

        \hline
        \textbf{Network size} & 416 & 384 & 352 & 320 & 288 & 256 & 224 & 192 & 160 & 128 & 96\\
        \hline
        \textbf{Foveal recall} & 35.20 & 35.27 & 35.03 & 34.30 & 34.83 & 33.98 & 34.30 & 35.33 & 33.62 & 32.38 & 23.32\\
        \textbf{Normal recall} & 34.57 & 33.75 & 33.15 & 31.89 & 30.82 & 29.29 & 27.60 & 25.07 & 20.30 & 17.33 & 6.42 \\
        \hline
        \hline
        \textbf{Foveal precision} & 25.09 & 25.66 & 25.51 & 25.36 & 26.47 & 25.95 & 27.58 & 29.43 & 30.70 & 34.70 & 30.80\\
        \textbf{Normal precision} & 24.45 & 24.09 & 23.85 & 23.14 & 22.42 & 21.72 & 20.19 & 18.71 & 17.41 & 15.15 & 8.31 \\
        \hline
        \hline
        \textbf{20 object frame-rate} & 3.59 & 4.29 & 4.75 & 5.21 & 5.76 & 8.52 & 9.40 & 10.76 & 12.14 & 15.24 & 16.65\\
        \hline
        \textbf{80 object frame-rate} & 3.31 & 4.02 & 4.53 & 4.87 & 5.34 & 8.04 & 8.91 & 10.36 & 11.70 & 14.63 & 15.75\\
        \hline

        \end{tabular}
    \end{center}
\caption{Comparison of the precision and recall (\%) for only the object centered in the fovea, with image and network at varying resolutions, from 416 to 96 (per side), along with the frame-rate average on a Jetson TX2 board.}
\label{table:retResults}
\end{table*}

%The main focus here is to compare the efficiency of running the neural network on the smallest scale that the foveal training allows, and therefore obtaining a considerable speed-up, for use on embedded systems. Table~\ref{table:fovTable} synthesizes the trade-off between recall and speed-up on a Jetson TX2 board. With normally downsampled images as input, the recall roughly halves, while the processing speed per frame can be more than quadrupled.

%\begin{table}
%    \begin{center}
%        \begin{tabular}{|c|c|c|c|c|c|c|}       %p{2cm}

%        \hline
%        \textbf{Size} & 416 & 192 & 160 & 128 & 96\\
%        \hline
%        \textbf{Foveal} & 1 & 1.003 & 0.955 & 0.919 & 0.663\\
%        \textbf{Normal} & 0.982 & 0.712 & 0.577 & 0.492 & 0.182 \\
%        \hline
%        \hline
%        \textbf{Speed-up} & 1 & 2.997 & 3.381 & 4.245 & 4.637\\
%        \hline

%        \end{tabular}
%    \end{center}
%\caption{Recall for objects in the fovea, proportional to the performance at 416x416, with foveal and normal downsamples, along with the speed-up for detection of 20 objects on a Jetson TX2.}
%\label{table:fovTable}
%\end{table}

\begin{table}       %[!htb]
    \begin{minipage}{.5\linewidth}
      
      \centering
        \begin{tabular}{|c|c|c|c|c|c|c|}
            
        \hline
        \textbf{Size} & 416 & 192 & 160 & 128 & 96\\
        \hline
        \textbf{Foveal} & 1 & 1.003 & 0.955 & 0.919 & 0.663\\
        \textbf{Normal} & 0.982 & 0.712 & 0.577 & 0.492 & 0.182 \\
        \hline
        \hline
        \textbf{Speed-up} & 1 & 2.997 & 3.381 & 4.245 & 4.637\\
        \hline
        \end{tabular}

      \label{table:fovTable}
      
    \end{minipage}%
    \begin{minipage}{.5\linewidth}
      \centering
        \begin{tabular}{|c|c|c|c|c|c|c|}
        
        \hline
        \textbf{Size} & 416 & 192 & 160 & 128 & 96\\
        \hline
        \textbf{Foveal} & 1 & 0.591 & 0.494 & 0.343 & 0.200\\
        \textbf{Normal} & 1.034 & 0.739 & 0.543 & 0.453 & 0.157 \\
        \hline

        \end{tabular}
        
       \label{table:perTable}
    \end{minipage} 
    \\
     \caption{\textbf{(Left)} Recall for objects in the fovea, proportional to the performance at 416x416, with foveal and normal downsamples, along with the speed-up for detection of 20 objects on a Jetson TX2. \textbf{(Right)} Recall comparison for objects in the periphery, proportional to the performance at 416x416.}
    
\end{table}

%While studying the object in the fovea is the main task for a given time, and for which most computing power is assigned, it is equally important to locate the most promising regions of interest to test on subsequent frames. The overall performance is plotted in the bottom right corner of Fig.~\ref{Coco20RetrainedA} and resumed in Table~\ref{table:perTable}.

%\begin{table}
%    \begin{center}
%        \begin{tabular}{|c|c|c|c|c|c|c|}

%        \hline
%        \textbf{Size} & 416 & 192 & 160 & 128 & 96\\
%        \hline
%        \textbf{Foveal} & 1 & 0.591 & 0.494 & 0.343 & 0.200\\
%        \textbf{Normal} & 1.034 & 0.739 & 0.543 & 0.453 & 0.157 \\
%        \hline

%        \end{tabular}
%    \end{center}
%\caption{Recall comparison for objects in the periphery, proportional to the performance at 416x416.}
%\label{table:perTable}
%\end{table}

%\pagebreak

\subsection{Comparison between Faster R-CNN and YOLO un-retrained}

\begin{figure}  %[htbp]
\centerline{\includegraphics[width=1.0\textwidth]{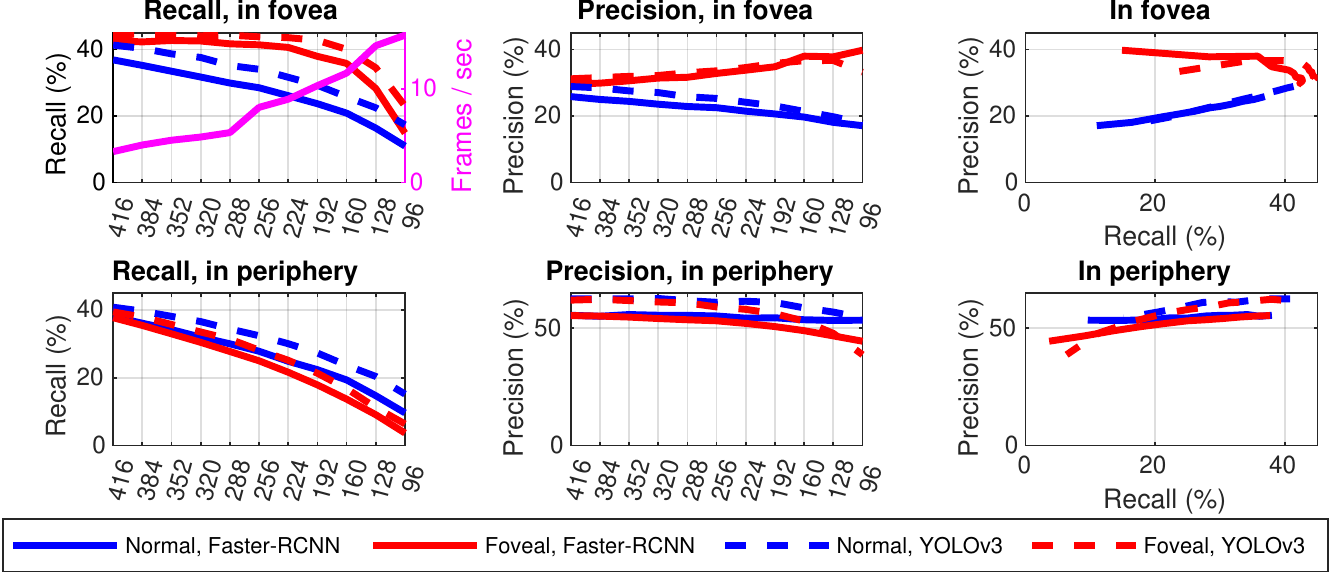}}
\caption{Comparison of performance changes using the un-retrained YOLOv3 and un-retrained Faster R-CNN neural networks as they were trained by their initial contributors on 80 object classes. Note that frames per second in the top panel is for YOLOv3 only.}
\label{Coco20noRetrainA}
\end{figure}

\begin{figure}  %[ht]  %[htbp]
\centerline{\includegraphics[width=1.0\textwidth]{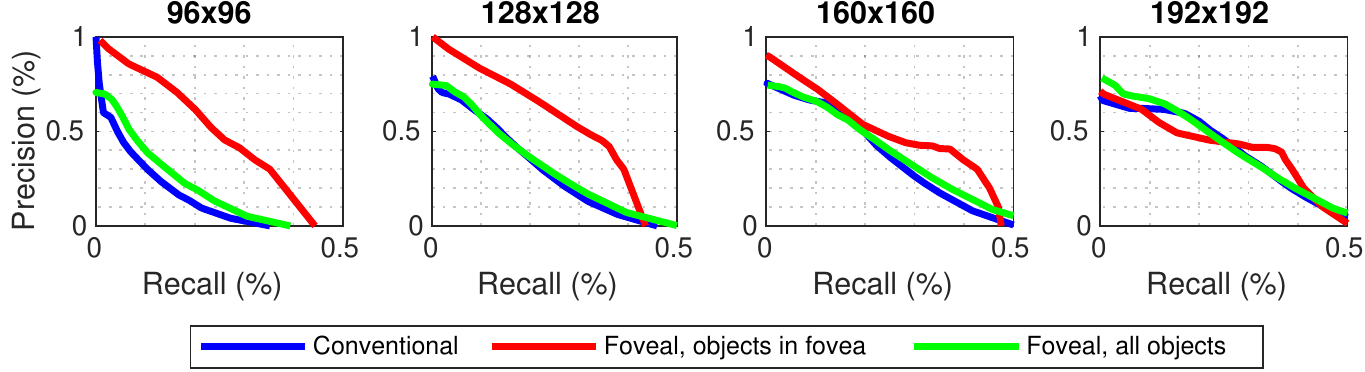}}
\caption{Precision and recall performance curves for the network at small resolutions (96x96, 128x128, 160x160 and 192x192). The foveal advantage is much more evident for the smaller networks, where the speed-up is also larger. In all cases, the performance is very similar between the normal downsample and the average of all objects present in the foveated image. For the larger networks, the prospect of detecting objects in the periphery increases, which affects the precision measurements when only the foveal object is considered as a true positive.}
\label{Recall_precision}
\end{figure}

To corroborate the previous observations, Faster-RCNN was also tested with the foveated images and uniformly sampled images at varying sizes. Due to the lengthy training process, retraining was avoided here for Faster-RCNN, and to provide a consistent comparison an un-retrainined version of YOLOv3 was also used in this section. Both networks were used with all 80 object classes from their original versions. The behaviour was remarkably similar to that obtained in the previous sections, for both YOLO and Faster R-CNN. (Fig. \ref{Coco20noRetrainA}). A $4\times$ speed-up in frame rate was still observed for YOLOv3, from 3.31 FPS to 14.63 FPS at $416\times416$ and $128 \times 128$ respectively (Fig. \ref{Coco20noRetrainA}). This serves as some confirmation that the approach of using foveated images with reduced size is beneficial to wider CNN designs used in object detection and recognition. These results also provide evidence that the advantages of foveation in object detection are not simply due to an effect of detecting image distortion due to the foveated transform itself.

\begin{figure}  %[htbp]
\centerline{\includegraphics[width=1.0\textwidth]{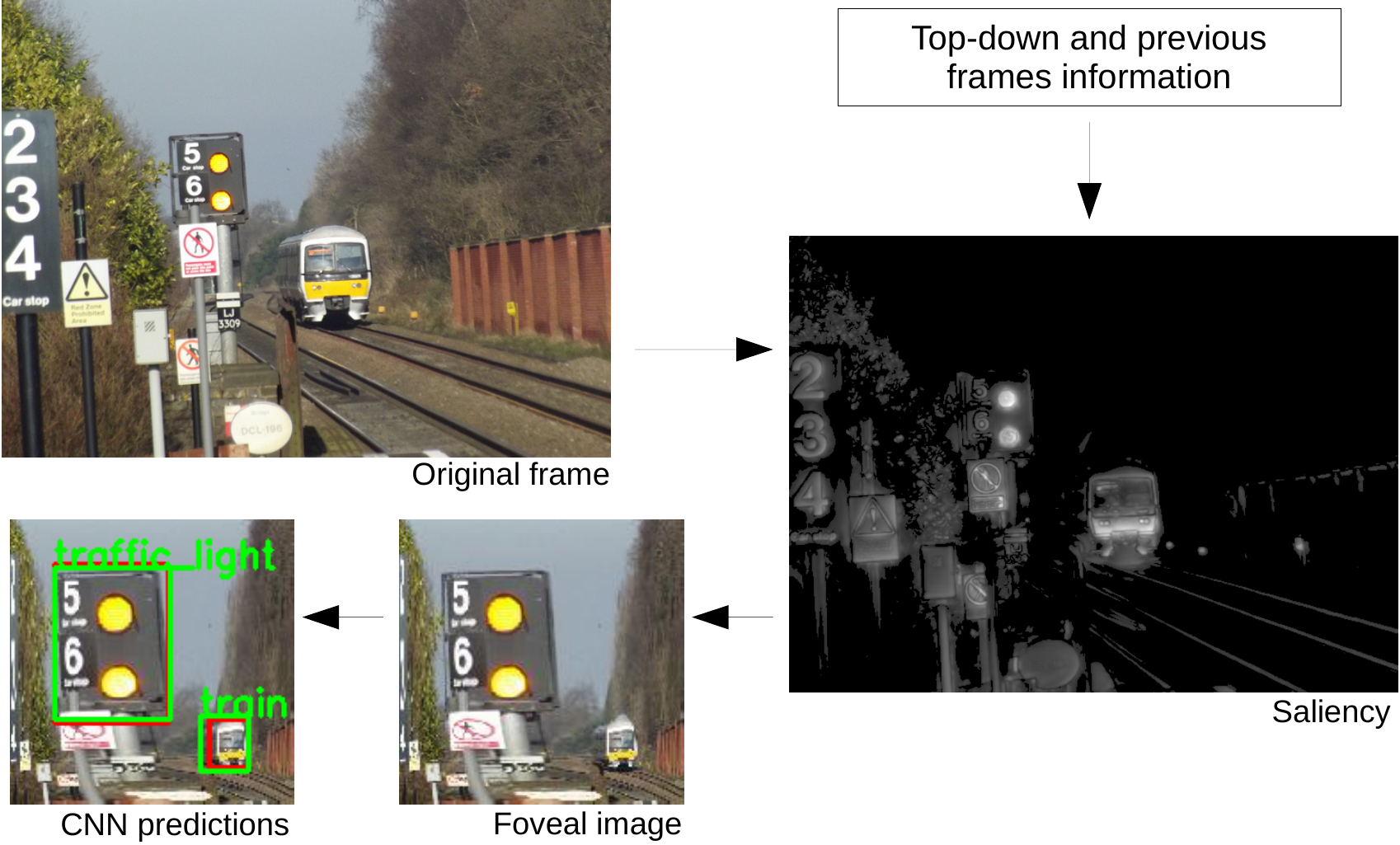}}
\caption{Example of an image from the Coco dataset run through the Vocus2 bottom-up saliency algorithm \cite{frintrop2015traditional}, then taking the most salient point as the center of the fovea. In this example the original image is 640x480 pixels, which in this case is downsampled to 160x160 pixels.}
\label{stoplightA}
\end{figure}

\section{Discussion}

\subsection{Implication of results}

The motivation for this study was to make object detection and recognition with CNNs more efficient for embedded GPU systems. The aim was
to investigate quantitatively how detection, recognition and processing speed in a CNN were affected by reducing image size using a foveated transformation. The results demonstrated that images could be reduced in size from $416\times 416$ to $128\times128$ pixels, with only a small decrease, 8.0\%, in recall using foveated sampling. A limitation of the approach was the decrease in object detection and recognition in the periphery, which was to be expected given the downsampling of pixels. 
%However, it is not clear from this study whether, or how, this might adversely affect an active vision system that re-directs the fovea to salient targets, which will be important to assess in future work. 

The key benefit observed here was the processing speed-up for reduced size images, specifically a $4\times$ speed-up with $128\times128$ pixel images. The increase in processing speed observed here is advantageous for future embedded systems: in the short term embedded systems with limited GPU processing power can more readily exploit the latest advanced algorithms, whilst in the long term as GPUs advance, less resource will be needed for object detection and recognition, maximising resources and energy efficiency.

\subsection{Future work}

The foveation method investigated here, in practice, would be part of a wider active vision system, incorporating saliency to re-direct the fovea to objects of interest. This is a key area to develop in future work. In order to illustrate how visual saliency and foveated object detection-recognition might function in practice, we have demonstrated the method developed here in combination with a well established bottom-up saliency algorithm \cite{frintrop2015traditional} (Fig. \ref{stoplightA}). 

Several methods have been developed to build-in saliency into the structure of the CNN itself \cite{zhang2018new}. It may be possible, therefore, to realise improved computational efficiency if saliency and object detection-recognition layers in the CNN are shared. This would be an interesting area of future work.

{\small
\bibliographystyle{splncs04}
\bibliography{Foveabib}
}

\end{document}